\documentclass[12pt]{llncs}
\usepackage{epsf,epsfig,makeidx}
\def\qlr{\mbox{\em QlR}}
\def\olhp{\mbox{\em olhp}}
\def\clhp{\mbox{\em clhp}}
\def\orhp{\mbox{\em orhp}}
\def\crhp{\mbox{\em crhp}}
\def\iaff{\mbox{\em iff}}
\def\lbound{\langle ^{\imath}}
\def\rbound{\rangle ^{\jmath}}
\def\tcsp{\mbox{\em TCSP}}
\def\tcsps{\mbox{\em TCSPs}}
\def\stp{\mbox{\em STP}}
\def\stps{\mbox{\em STPs}}
\def\dmp{\mbox{\em DMP}}
\def\scsp{\mbox{\em SCSP}}
\def\scsps{\mbox{\em SCSPs}}
\def\ssp{\mbox{\em SSP}}
\def\bsp{\mbox{\em BSP}}
\def\bsps{\mbox{\em BSPs}}
\def\ssps{\mbox{\em SSPs}}
\def\rcc8{\mbox{RCC-8}}
\def\wrt{\mbox{w.r.t.}}

\def\alcd{{\cal ALC}({\cal D})}
\def\cdalg{{\cal CDA}}
\def\cdalgcs{{\cal CDA}_{cs}}
\def\cdalgpb{{\cal CDA}_{pb}}
\def\binmat{{\cal B}}

\def\rel-alg{{\cal U}}
\def\north{{\mbox{\em No}}}
\def\northeast{{\mbox{\em NE}}}
\def\east{{\mbox{\em Ea}}}
\def\southeast{{\mbox{\em SE}}}
\def\south{{\mbox{\em So}}}
\def\southwest{{\mbox{\em SW}}}
\def\west{{\mbox{\em We}}}
\def\northwest{{\mbox{\em NW}}}
\def\equal{{\mbox{\em Eq}}}
\def\northcs{{\mbox{\em No}_{cs}}}
\def\northeastcs{{\mbox{\em NE}_{cs}}}
\def\eastcs{{\mbox{\em Ea}_{cs}}}
\def\southeastcs{{\mbox{\em SE}_{cs}}}
\def\southcs{{\mbox{\em So}_{cs}}}
\def\southwestcs{{\mbox{\em SW}_{cs}}}
\def\westcs{{\mbox{\em We}_{cs}}}
\def\northwestcs{{\mbox{\em NW}_{cs}}}
\def\equalcs{{\mbox{\em Eq}_{cs}}}
\def\northpb{{\mbox{\em No}_{pb}}}
\def\northeastpb{{\mbox{\em NE}_{pb}}}
\def\eastpb{{\mbox{\em Ea}_{pb}}}
\def\southeastpb{{\mbox{\em SE}_{pb}}}
\def\southpb{{\mbox{\em So}_{pb}}}
\def\southwestpb{{\mbox{\em SW}_{pb}}}
\def\westpb{{\mbox{\em We}_{pb}}}
\def\northwestpb{{\mbox{\em NW}_{pb}}}
\def\equalpb{{\mbox{\em Eq}_{pb}}}
\def\BBR{{\rm I\!R}}

\begin{document}
\mainmatter
\title{Integrating existing cone-shaped and projection-based cardinal direction
	relations and a $\tcsp$\thanks{$\tcsps$ stands for Temporal Constraint Satisfaction Problems, a well-known constraint-based temporal framework \cite{Dechter91a}.}-like decidable generalisation\thanks{This work was supported partly by
	the EU project ``{\em Cognitive Vision systems}" (CogVis), under grant
	{\em CogVis IST 2000-29375}.}}
\titlerunning{Spatial CSPs}
\author{Amar Isli}
\authorrunning{Isli}
\institute{Fachbereich Informatik, Universit\"at Hamburg,\\
           Vogt-K\"olln-Strasse 30, D-22527 Hamburg, Germany\\
	   \email{isli@informatik.uni-hamburg.de}
}

\maketitle

\begin{flushright}
\begin{tiny}
Conciliating qualitative reasoning and quantitative reasoning in KR\&R systems:\\
a way to systems
	representationally more flexible,
	cognitively more plausible,
	and, 	computationally, with the advantage of having the choice between
		a purely-quantitative and a
		qualitative-computations-first behaviours.
\end{tiny}
\end{flushright}

\begin{abstract}
Integrating different knowledge representation languages is clearly
an important topic. This allows, for instance, for a unified
representation of knowledge coming from different sources, each
source using one of the integrated languages for its knowledge
representation. This is of special importance
for QSR\footnote{Qualitative Spatial Reasoning.} languages, for such
a language makes only a finite number of distinctions: integrating
QSR languages may be looked at as an answer to the well-known
poverty conjecture. With these considerations in mind, we consider
the integration of Frank's cone-shaped and projection-based calculi
of cardinal direction relations, well-known in QSR.
The more general, integrating language we consider is based on
convex constraints of the qualitative form $r(x,y)$, $r$ being a
cone-shaped or projection-based cardinal direction atomic relation,
or of the quantitative form $(\alpha ,\beta )(x,y)$, with
$\alpha ,\beta\in [0,2\pi )$ and $(\beta -\alpha )\in [0,\pi ]$:
the meaning of the quantitative constraint, in particular, is
that point $x$ belongs to the (convex) cone-shaped area rooted at $y$,
and bounded by angles $\alpha$ and $\beta$.
The general form of a constraint is a disjunction of the form
$[r_1\vee\cdots\vee r_{n_1}\vee (\alpha _1,\beta _1)\vee\cdots\vee (\alpha _{n_2},\beta _{n_2})](x,y)$,
with $r_i(x,y)$, $i=1\ldots n_1$, and $(\alpha _i,\beta _i)(x,y)$,
$i=1\ldots n_2$, being convex constraints as described above: the
meaning of such a general constraint is that, for some $i=1\ldots n_1$,
$r_i(x,y)$ holds, or, for some $i=1\ldots n_2$,
$(\alpha _i,\beta _i)(x,y)$ holds. A conjunction of such general
constraints is a $\tcsp$-like CSP, which we will refer to as an
$\scsp$ (Spatial Constraint Satisfaction Problem). An effective
solution search algorithm for an $\scsp$ will be described, which
uses (1) constraint propagation, based on a composition operation to
be defined, as the filtering method during the search,
and (2) the Simplex algorithm, guaranteeing completeness, at the
leaves of the search tree. The approach is particularly suited for
large-scale high-level vision, such as, e.g., satellite-like
surveillance of a geographic area.
\end{abstract}
\newtheorem{cor}{Corollary}   
\newtheorem{rem}{Remark}
\newtheorem{thr}{Theorem}
\newtheorem{df}{Definition}
\newtheorem{fact}{Fact}
\newtheorem{lem}{Lemma}
\newtheorem{ex}{Example}
\section{Introduction}\label{sect1}
Knowledge representation (KR) systems allowing for the representation of both
qualitative knowledge and quantitative knowledge are more than needed by modern
applications, which, depending on the level of detail of the knowledge to be
represented, may feel happy with a high-level, qualitative language, or need to use a
low-level, quantitative language. Qualitative languages suffer from what Forbus et al.
\cite{Forbus91a} refer to as the poverty conjecture (which corresponds more or less to Habel's \cite{Habel95a}
argument that such languages suffer from not having ``the ability to refine discrete
structures if necessary''), but have the advantage of bahaving computationally better.
On the other hand, quantitative languages do not suffer from the poverty conjecture,
but have a slow computatinal behaviour. Thus, such a KR system will feel happier when
the knowledge at hand can be represented in a purely qualitative way, for it can then get rid of heavy numeric calculations, and restrict its computations to symbols manipulation, consisting, in the case of constraint-based languages in the style of the Region-Connection Calculus $\rcc8$ \cite{Randell92a}, mainly in computing a closure under a composition table.

An important question raised by the above discussion is clearly how to augment the chances of a qualitative/quantitative KR system to remain at the qualitative level. Consider, for instance, QSR constraint-based, $\rcc8$-like languages. Given the poverty conjecture, which corresponds to the fact that such a language can make only a finite number of distinctions, reflected by the number of its atomic relations, one way of answering the question could be to integrate more than one QSR language within the same KR system. The knowledge at hand is then handled in a quantitative way only in the extreme case when it can be represented by none of the QSR languages which the system integrates.

One way for a KR system, such as described above, to reason about its
knowledge is to start with reasoning about the qualitative part of
the knowledge, which decomposes, say, into n components, one for
each of the QSR languages the system integrates. For $\rcc8$-like
languages, this can be done using a constraint propagation algorithm
such as the one in \cite{Allen83b}. If in either of the n components,
an inconsistency has been detected, then the whole knowledge has been
detected to be inconsistent without the need of going into low-level
details. If no inconsistency has been detected at the high,
qualitative level, then the whole knowledge needs translation into
the unifying quantitative language, and be processed in a purely
quantitative way. But even when the high-level, qualitative
computations fail to detect any inconsistency, they still potentially
help the task of the low-level, purely quantitative computations.
The situation can be compared to standard search algorithms in CSPs,
where a local-consistency preprocessing is applied to the whole
knowledge to potentially reduce the search space, and eventually detect
the knowledge inconsistency, before the actual search for a solution
starts.

With the above
considerations in mind, we consider the integration of Frank's
cone-shaped and projection-based calculi of cardinal direction
relations \cite{Frank92b}, well-known in QSR. A complete decision procedure for the
projection-based calculus is known from Ligozat's work \cite{Ligozat98a}. For the other
calculus, based on a uniform 8-sector partition of the plane, making
it more flexible and cognitively more plausible, no
such procedure is known. For each of the two calculi, the region of
the plane associated with each of the atomic relations is convex,
and given by the intersection of two half-planes. As a
consequence, each such relation can be equivalently written as a
conjunction of linear inequalities on variables consisting of the
coordinates of the relation's arguments. We
consider a more general, qualitative/quantitative language, which, at
the basic level, expresses convex constraints of the form $r(x,y)$, where $r$ is a cone-shaped
or projection-based atomic relation of cardinal directions, or of the
form $(\alpha ,\beta )(x,y)$, with $\alpha ,\beta\in [0,2\pi )$ and
$(\beta -\alpha )\in [0,\pi ]$: the meaning of $(\alpha ,\beta )(x,y)$, in particular, is
that point $x$ belongs to the (convex) cone-shaped area rooted at $y$,
and bounded by angles $\alpha$ and $\beta$. We refer to such
constraints as basic constraints: qualitative basic constraint in the
first case, and quantitative basic constraint in the second. A conjunction of
basic constraints can be solved by first applying constraint
propagation, based on a composition operation to be defined, which is
basically the spatial counterpart of composition of two TCSP
constraints \cite{Dechter91a}. If the propagation detects no
inconsisteny then the knowledge is translated into a system of
linear inequalities, and solved with the well-known Simplex
algorithm. The preprocessing of the qualitative component of the
knowledge can be done with a constraint propagation algorithm such
as the one in \cite{Allen83b}, and needs the composition tables of
the cardinal direction calculi, which can be found in
\cite{Frank92b}.

To summarise, given combined qualitative/quantitative conjunctive knowledge,
expressed as a conjunction of basic constraints, the reasoning methodology we propose works in
three steps:
\begin{enumerate}
  \item First apply (qualitative) constraint propagation \cite{Allen83b} to each
	of the qualitative components of the knowledge.
  \item If no inconsistency has been detected by the previous step,
	then translate the qualitative knowledge into quantitative
	knowledge, so that the whole knowledge gets expressed in
	the unifying quantitative language; then apply (quantitative) constraint
	propagation to the whole, based on a composition operation
	to be defined later.
  \item If no inconsistency has been detected by the previous step,
	then translate the knowledge into a conjunction of linear
	inequalities, and apply the complete Simplex procedure to
	decide whether the knowledge is consistent.
\end{enumerate}
The general form of a constraint is $(s_1\vee\cdots\vee s_n)(x,y)$, which we
also represent as $\{s_1,\ldots ,s_n\}(x,y)$, where $s_i(x,y)$, for all
$i\in\{1,\ldots ,n\}$, is a basic constraint, either qualitative or
quantitative. The meaning of such a general constraint is that, either
of the $n$ basic constraints is satisfied, i.e.,
$s_1(x,y)\vee\cdots\vee s_n(x,y)$. A general constraint is qualitative if
it is the disjunction of qualitative basic constraints of one type,
cone-shaped or projection-based; it is
quantitative otherwise. The language can be looked at as the spatial
conterpart of Dechetr et al.'s TCSPs \cite{Dechter91a}: the domain
of a TCSP variable is $\BBR$, symbolising continuous time, whereas the
domain of an SCSP variable is the cross product $\BBR\times\BBR$,
symbolising the continuous 2-dimensional space.

The reasoning module of our KR system involves thus two known
techniques: constraint propagation, based on composition of two
basic constraints, and the Simplex algorithm. If both basic
constraints are qualitative, and both cone-shaped or both
projection-based, then their composition is given by existing
composition tables \cite{Frank92b}. Otherwise, the basic
constraints are considered as quantitative, and their composition
is computed in a way to be defined later (similar to composition
of two convex TCSP constraints \cite{Dechter91a}).

Some emphasis on our approach to knowledge representation is needed.
Researchers working on purely
quantitative languages use arguments such as the poverty conjecture in
\cite{Forbus91a} to criticise qualitative reasoning in general, and QSR in
particular. On the other hand , QlR\footnote{We use QlR and QnR as
shorthands for Qualitative Reasoning and for Quantitative Reasoning,
resepectively.} researchers argue that quantitative
reasoning goes often too much into unnecessary details, which is reflected by
ideas such as ``make only as many distinctions as necessary''
\cite{Cohn97b}, borrowed to na\"{\i}ve physics
\cite{Hayes85b}. Our approach is a conciliating one, and is meant to satisfy both
tendencies. It consists of combining $\qlr$ languages known to be
sufficient for a large number of applications, with a subsuming
quantitative language. The number of $\qlr$ languages may be, as in the
present work, more than just one, to allow potential applications high
chances to remain at the high-level, qualitative languages for their
knowledge representation. $\qlr$ researchers are satisfied since they have
the possibility of using only the qualitative part of the language. On
the other hand, if an application needs more expressiveness than is
allowed by any of the $\qlr$ sublanguages, then the unifying quantitative
language is there to satisfy it.

Current research shows clearly the importance of developing decidable
constraint-based spatial languages: specialising an $\alcd$-like
Description Logic (DL) \cite{Baader91a}, so that the roles are
temporal immediate-successor (accessibility) relations, and the
concrete domain is generated by a decidable constraint-based spatial
language, such as an $\rcc8$-like qualitative spatial RA
\cite{Randell92a}, or a combined qualitative/quantitative language
such as the one to be described in this paper, leads to a
computationally well-behaving family of languages for spatial change
in general, and for motion of spatial scenes in particular:
\begin{enumerate}
  \item Deciding satisfiability of an $\alcd$ concept with respect
	to ($\wrt$) a cyclic TBox is, in general, undecidable
	(see, for instance, \cite{Lutz03a}).
  \item In the case of the spatio-temporalisation, however, if we
	use what is called weakly cyclic TBoxes in \cite{Isli02d},
	then satisfiability of a concept $\wrt$ such a TBox is
	decidable. The axioms of a weakly cyclic TBox capture the
	properties of modal temporal operators. The reader is
	referred to \cite{Isli02d} for details.
\end{enumerate}
Spatio-temporal theories such as the ones defined in \cite{Isli02d}
can be seen as single-ontology spatio-temporal theories, in the sense
that the concrete domain represents only one type of spatial knowledge 
(e.g., \mbox{RCC-8} relations if the concrete domain is generated by
\mbox{RCC-8}). The calculus to be defined can, of course, generate
such a single-ontology spatio-temporal theory; but with the
disadvantage that the concrete domain would be heterogeneous, in the
sense that it would group together two qualitative languages and a
unifying quantitative language, which some applications might not find
clean. We could, instead, use a 3-ontology spatio-temporal theory: two
ontologies of the theory generated by the cone-shaped and the
projection-based calculi of cardinal direction relations, the third
ontology by the unifying quantitative language.
\section{Constraint satisfaction problems}\label{csps}
A constraint satisfaction problem (CSP) of order $n$ consists of:
\begin{enumerate}
  \item a finite set of $n$ variables, $x_1,\ldots ,x_n$;
  \item a set $U$ (called the universe of the problem); and
  \item a set of constraints on values from $U$ which may be assigned to the
    variables.
\end{enumerate}
An $m$-ary constraint is of the form $R(x_{i_1},\cdots ,x_{i_m})$, and asserts
that the values $a_{i_1},\ldots ,a_{i_m}$ assigned to the variables $x_{i_1},\ldots ,x_{i_m}$, respectively,
are so that the $m$-tuple $(a_{i_1},\ldots ,a_{i_m})$ belongs the $m$-ary relation $R$ (an $m$-ary relation over the
universe $U$ is any subset of $U^m$). An $m$-ary CSP is one of which the
constraints are $m$-ary constraints. We will be concerned exclusively
with binary CSPs.

For any two binary relations $R$ and $S$, $R\cap S$ is the
intersection of $R$ and $S$,
$R\cup S$ is the union of $R$ and $S$, $R\circ S$ is the composition of $R$ and $S$, 
and $R^\smile$ is the converse of $R$; these are defined as follows:
\begin{center}
$
\begin{array}{lll}
R\cap S      &=&\{(a,b):(a,b)\in R\mbox{ and }(a,b)\in S\},\\
R\cup S      &=&\{(a,b):(a,b)\in R\mbox{ or }(a,b)\in S\},\\
R\circ S   &=&\{(a,b):\mbox{for some }c,(a,c)\in R\mbox{ and }(c,b)\in S\},\\
R^\smile     &=&\{(a,b):(b,a)\in R\}.
\end{array}
$
\end{center}
Three special binary relations over a universe $U$ are the empty
relation $\emptyset$ which contains
no pairs at all, the identity relation ${\cal I}_U^b=\{(a,a):a\in U\}$, and the universal
relation $\top _U^b=U\times U$.

Composition and converse for binary relations were introduced by De Morgan
\cite{DeMorgan1864a,DeMorgan66a}.
\subsection{Constraint matrices}
A binary constraint matrix of order $n$ over $U$ is an $n\times n$-matrix, say $\binmat$,
of binary relations over $U$ verifying the following:
\begin{center}
$
\begin{array}{ll}
(\forall i\leq n)(\binmat _{ii}\subseteq {\cal I}_U^b)   &\mbox{(the diagonal property)},\\
(\forall i,j\leq n)(\binmat _{ij}=(\binmat _{ji})^\smile )&\mbox{(the converse property)}.
\end{array}
$
\end{center}
A binary CSP $P$ of order $n$ over a universe $U$ can be associated with the
following binary constraint matrix, denoted $\binmat ^P$:
\begin{enumerate}
  \item Initialise all entries to the universal relation:
    $(\forall i,j\leq n)((\binmat ^P)_{ij}\leftarrow \top _U^b)$
  \item Initialise the diagonal elements to the identity relation:\\
    $(\forall i\leq n)((\binmat ^P)_{ii}\leftarrow {\cal I}_U^b)$
  \item For all pairs $(x_i,x_j)$ of variables on which a
    constraint $(x_i,x_j)\in R$ is specified:
    $(\binmat ^P)_{ij}\leftarrow (\binmat ^P)_{ij}\cap R,(\binmat ^P)_{ji}\leftarrow ((\binmat ^P)_{ij})^\smile$.
\end{enumerate}
We make the assumption that, unless explicitly specified otherwise, a CSP is
given as a constraint matrix.
\subsection{Strong $k$-consistency, refinement}
Let $P$ be a CSP of order $n$, $V$ its set of variables and $U$ its universe.
An instantiation of $P$ is any $n$-tuple $(a_1,a_2,\ldots ,a_n)$ of $U^n$,
representing an assignment of a value to each variable.  A consistent instantiation
is an instantiation $(a_1,a_2,\ldots ,a_n)$ which is a solution:
$(\forall i,j\leq n)((a_i,a_j)\in (\binmat ^P)_{ij})$.
$P$ is consistent if it has at least one solution; it is inconsistent otherwise. The
consistency problem of $P$ is the problem of verifying whether $P$ is consistent.

Let $V'=\{x_{i_1},\ldots ,x_{i_j}\}$ be a subset of $V$. The sub-CSP of $P$ generated
by $V'$, denoted $P_{|V'}$, is the CSP with $V'$ as the set of variables, and whose constraint
matrix is obtained by projecting the constraint matrix of $P$ onto $V'$:
$(\forall k,l\leq j)((\binmat ^{P_{|V'}})_{kl}=(\binmat ^P)_{i_ki_l})$.
$P$ is $k$-consistent \cite{Freuder78a,Freuder82a} (see also \cite{Cooper89a}) if for any subset $V'$ of $V$
containing $k-1$ variables, and for any variable $X\in V$, every solution to
$P_{|V'}$ can be extended to a solution to $P_{|V'\cup\{X\}}$. $P$ is strongly
$k$-consistent if it is $j$-consistent, for all $j\leq k$.

$1$-consistency, $2$-consistency and $3$-consistency correspond to node-consistency,
arc-consistency and path-consistency, respectively \cite{Mackworth77a,Montanari74a}.
Strong $n$-consistency of $P$ corresponds to what is called global consistency in
\cite{Dechter92a}. Global consistency facilitates the important task of searching
for a solution, which can be done, when the property is met, without backtracking
\cite{Freuder82a}.

A refinement of $P$ is a CSP $P'$ with the same set of variables, and such that:
$(\forall i,j)((\binmat ^{P'})_{ij}\subseteq (\binmat ^P)_{ij})$.
\section{Frank's calculi of cardinal direction relations}
\begin{figure}[t]
\centerline{\epsfxsize=11cm\epsfysize=4.75cm\epsffile{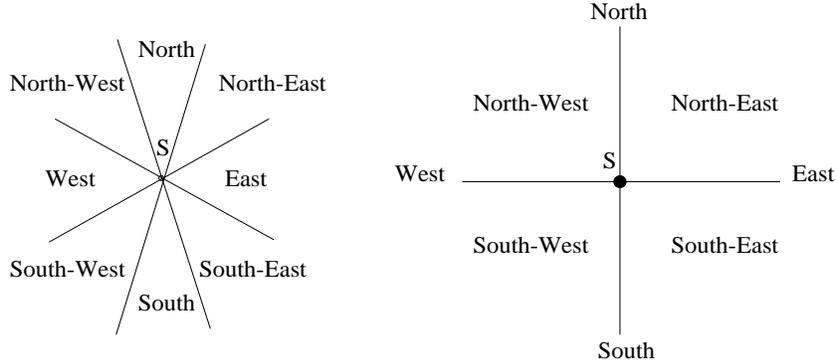}}
\caption{Frank's cone-shaped (left) and projection-based (right) models of
cardinal directions.}\label{sectors}
\end{figure}
Frank's models of cardinal directions in 2D \cite{Frank92b} are
illustrated in Figure \ref{sectors}. They use a partition of the plane
into regions determined by lines passing through a reference object,
say $S$. Depending on the region a point $P$ belongs to, we have
$\north (P,S)$, $\northeast (P,S)$, $\east (P,S)$, $\southeast (P,S)$,
$\south (P,S)$, $\southwest (P,S)$, $\west (P,S)$, $\northwest (P,S)$,
or $\equal (P,S)$, corresponding, respectively, to the position of $P$
relative to $S$ being $\mbox{\em north}$, $\mbox{\em north-east}$,
$\mbox{\em east}$, $\mbox{\em south-east}$, $\mbox{\em south}$,
$\mbox{\em south-west}$, $\mbox{\em west}$, $\mbox{\em north-west}$,
or $\mbox{\em equal}$. Each of the two models can thus be seen as a
binary Relation Algebra (RA), with nine atoms. Both use a global
\mbox{horizontal/vertical}, {\em left-right/bottom-up} reference frame,
which we suppose to be a Cartesian coordinate system $(O,x'x,y'y)$. The
coordinate system so chosen clearly verifies the fact that, on the one
hand, the \mbox{$x$-axis} $x'x$ is parallel to, and has the same
orientation as the \mbox{West-East} directed line of Frank's
\mbox{projection-based} model, and, on the other hand, the
\mbox{$y$-axis} $y'y$ is parallel to, and has the same
orientation as the \mbox{South-North} directed line of the same model ---$O$ is the intersection of the \mbox{$x$-} and \mbox{$y$-axes}.

To differentiate between the two models, we use the
underscore {\em cs} for the cone-shaped model, and the underscore
\mbox{pb} for the projection-based model. Thus, from now on,
(1) we refer to the cone-shaped model as $\cdalgcs$, and to the
projection-based model as $\cdalgpb$; and
(2) we denote the atoms of $\cdalgcs$ as $\northcs$, $\northeastcs$, $\eastcs$,
$\southeastcs$, $\southcs$, $\southwestcs$, $\westcs$, $\northwestcs$, and
$\equalcs$, and the atoms of $\cdalgpb$ as $\northpb$, $\northeastpb$,
$\eastpb$, $\southeastpb$, $\southpb$, $\southwestpb$, $\westpb$, $\northwestpb$,
and $\equalpb$.

A $\cdalgcs$ (resp. $\cdalgpb$) relation is any
subset of the set of all $\cdalgcs$ (resp. $\cdalgpb$) atoms. A $\cdalgcs$ (resp.
$\cdalgpb$) relation is said to be atomic if it contains one single atom (a
singleton set); it is said to be the $\cdalgcs$ (resp. $\cdalgpb$) universal
relation if it contains all the $\cdalgcs$ (resp. $\cdalgpb$) atoms. When no
confusion raises, we may omit the brackets in the representation of an
atomic relation.
\subsection{CSPs of cardinal direction relations on 2D points}
We define a $\cdalgcs$-CSP (resp. $\cdalgpb$-CSP) as a CSP of which the
constraints are $\cdalgcs$ (resp. $\cdalgpb$) relations on pairs of the
variables. The universe of such a CSP is the set $\BBR ^2$ of 2D points.

A $\cdalgcs$-matrix (resp. $\cdalgpb$-matrix) of order $n$ is a binary constraint
matrix of order $n$ of which the entries are $\cdalgcs$ (resp. $\cdalgpb$)
relations. The constraint matrix associated with a $\cdalgcs$-CSP (resp.
$\cdalgpb$-CSP) is a $\cdalgcs$-matrix (resp. $\cdalgpb$-matrix).
A scenario of such a CSP is a refinement $P'$ such that all entries
of the constraint matrix of $P'$ are atomic relations.
A CSP of cardinal direction relations that does not include the empty
constraint, which indicates a trivial inconsistency,
is strongly $2$-consistent. A $\cdalg$-CSP is a CSP which is either a
$\cdalgcs$-CSP or a $\cdalgpb$-CSP. An atomic $\cdalg$-CSP is a $\cdalg$-CSP
which is its own unique scenario (i.e., of which all entries of the
constraint matrix are atomic relations).
\subsection{Solving a $\cdalg$-CSP}
A simple adaptation of Allen's constraint propagation algorithm \cite{Allen83b}
can be used to achieve path consistency (hence strong $3$-consistency)
for a CSP of cardinal direction relations, thanks to composition tables of the
calculi which can be found in \cite{Frank92b}. Applied to such a CSP,
say $P$, such an adaptation would repeat the following steps until either
stability is reached or the empty relation is detected (indicating
inconsistency):
\begin{enumerate}
  \item Consider a triple $(X_i,X_j,X_k)$ of variables verifying
        $(\binmat ^P)_{ij}\not\subseteq (\binmat ^P)_{ik}\circ (\binmat ^P)_{kj}$
  \item $(\binmat ^P)_{ij}\leftarrow (\binmat ^P)_{ij}\cap (\binmat ^P)_{ik}\circ (\binmat ^P)_{kj}$
  \item If $((\binmat ^P)_{ij}=\emptyset)$ then exit (the CSP is inconsistent).
\end{enumerate}
Path consistency is complete for atomic $\cdalgpb$-CSPs
\cite{Ligozat98a}. Given this, Ladkin and Reinefeld's solution search
algorithm \cite{Ladkin92a} can be used to search for a solution, if
any, or otherwise report inconsistency, of a general $\cdalgpb$-CSP.
However, no such result is known for atomic $\cdalgcs$-CSPs. But even
so, we still can apply the search algorithm in \cite{Ladkin92a} to
search for a path-consistent scenario of a $\cdalgcs$-CSP,
if such a refinement exists, or report inconsistency otherwise. The
main result of the present work implies that we can solve the
consistency problem of an atomic $\cdalgcs$-CSP, by first translating
it into a conjunction of linear inequalities on variables consisting
of the coordinates of the point-variables of the $\cdalgcs$-CSP. This
means that for a general $\cdalgcs$-CSP, we can use the search
algorithm in \cite{Ladkin92a} augmented with the Simplex algorithm to
decide its consistency problem. The basic idea is to apply the
algorithm in \cite{Ladkin92a} as it is, and, whenever it succeeds to
find a path-consistent scenario (the algorithm is then at the level of
a leaf of the search tree), check, using the Simplex algorithm,
whether that scenario is consistent, by translating it into a
conjunction of linear inequalities. If the conjunction of linear
inequalities is consistent then the corresponding scenario is
consistent, and is thus a consistent scenario of the input
$\cdalgcs$-CSP. If the conjunction is inconsistent, then the search
for a possible consistent scenario has to continue. This is
illustrated in Figure \ref{backtrack}.
\begin{figure}
\begin{enumerate}
  \item[] {\em Input:} A $\cdalgcs$-CSP $P$;
  \item[] {\em Output:} {\em true} if and only if $P$ is consistent;
  \item[]{\em function} consistent($P$);
  \item \hspace{0.3cm} PC($P$);
  \item \hspace{0.3cm} {\em if}($P$ contains the empty relation)return false;
  \item \hspace{0.3cm} {\em else}
  \item \hspace{0.6cm} {\em if}($P$ contains edges labelled with relations other than atoms)\{
  \item \hspace{0.9cm} choose such an edge, say $(X_i,X_j)$;
  \item \hspace{0.9cm} $R\leftarrow (\binmat ^P)_{ij}$; \% save before branching \%
  \item \hspace{0.9cm} for each atom $r$ in $R$\{
  \item \hspace{1.2cm} refine $(\binmat ^P)_{ij}$ to $r$ (i.e., $(\binmat ^P)_{ij}\leftarrow r$);
  \item \hspace{1.2cm} {\em if}(consistent($P$))return {\em true};
  \item \hspace{1.2cm} \}
  \item \hspace{0.9cm} $(\binmat ^P)_{ij}\leftarrow R$; \% restore before backtracking \%
  \item \hspace{0.9cm} return false;
  \item \hspace{0.9cm} \}
  \item \hspace{0.6cm} {\em else}\{\hspace{1.7cm}\% path-consistent scenario found: \%
  \item[] \hspace{3cm} \% translate into linear inequalities and solve \%
  \item \hspace{0.9cm} translate $P$ into a conjunction of linear inequalities, $C$;
  \item \hspace{0.9cm} {\em if}($C$ has solutions)return true;
  \item \hspace{0.9cm} {\em else} return false
  \item \hspace{0.9cm} \}
\end{enumerate}
\caption{A consistent scenario search function for $\cdalgcs$-CSPs.}\label{backtrack}
\end{figure}
\section{Temporal Constraint Satisfaction Problems ---$\tcsps$}
$\tcsps$ have been proposed in \cite{Dechter91a} as an extension of
(discrete) CSPs \cite{Mackworth77a,Montanari74a} to continuous variables.
\begin{definition}[$\tcsp$ \cite{Dechter91a}]
A $\tcsp$ consists of (1) a finite number of variables ranging over the
universe of time points; and (2) Dechter, Meiri and Pearl's
constraints (henceforth DMP constraints) on the variables.
\end{definition}
A $\dmp$ constraint is either unary or binary. A unary constraint has
the form $R(Y)$, and a binary constraint the form $R(X,Y)$, where $R$ is a subset of the set $\BBR$ of real
numbers, seen as a unary relation in the former case, and as a binary
relation in the latter case, and $X$ and $Y$ are variables ranging over
the universe of time points: the unary constraint $R(Y)$ is interpreted as $Y\in R$, and
the binary constraint $R(X,Y)$ as $(Y-X)\in R$. A unary constraint $R(Y)$ may 
be seen as a special binary constraint if we consider an origin of the 
World (time $0$), represented, say, by a variable $X_0$: $R(Y)$ is
then equivalent to $R(X_0,Y)$. Unless explicitly stated otherwise, we
assume, in the rest of the paper, that the constraints of a $\tcsp$
are all binary.
\begin{definition}[$\stp$ \cite{Dechter91a}]
An $\stp$ (Simple Temporal Problem) is a $\tcsp$ of which all the
constraints are convex, i.e., of the form $R(X,Y)$, $R$ being a convex 
subset of $\BBR$.
\end{definition}
The universal relation for $\tcsps$ in general, and for $\stps$ in
particular, is the relation consisting of the whole set $\BBR$ of real
numbers: the knowledge $(Y-X)\in\BBR$, expressed by the $\dmp$ constraint
$\BBR (X,Y)$, is equivalent to ``no knowledge''. The identity
relation is the (convex) set reducing to the singleton $\{0\}$: the
constraint $\{0\}(X,Y)$ ``forces" variables $X$ and $Y$ to be equal.
\section{A spatial counterpart of $\tcsps$: Spatial Constraint Satisfaction Problems ($\scsps$)}
\begin{figure}
\centerline{\epsfxsize=7cm\epsfysize=7cm\epsffile{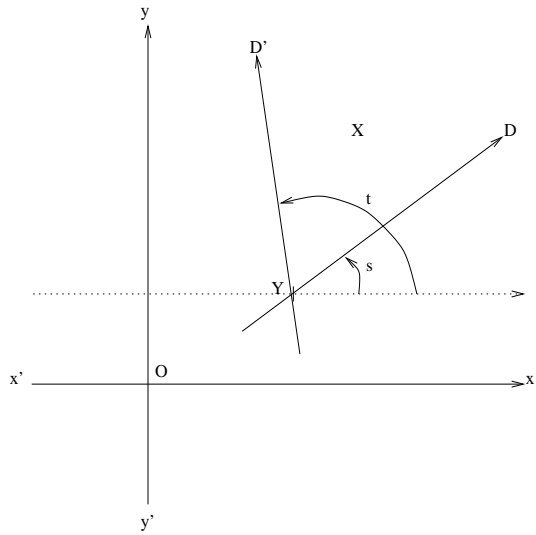}}
\caption{Graphical interpretation of the basic constraint $\lbound s,t\rbound (X,Y)$:
	Given $Y$, the set of points $X$ satisfying the constraint
	$\lbound s,t\rbound (X,Y)$ is the cone-shaped area centred at $Y$, whose
	lower bound (open if $\imath =0$, close otherwise) and upper bound (open if
	$\jmath =0$, close otherwise) are, respectively, the half-lines whose
	angular distances from the $x$-axis, with respect to anticlockwise
	orientation, are $s$ and $t$.}\label{basicconstraint}
\end{figure}

We now provide a spatial counterpart of $\tcsps$, which we refer to as $\scsps$
---Spatial Constraint Satisfaction Problems. The domain of an $\scsp$ variable is the cross
product $\BBR\times\BBR$, which we look at as the set of points of the 2-dimensional space. As
for a $\tcsp$, an $\scsp$ will have unary constraints and binary constraints, and unary
constraints can be interpreted as special binary constraints by choosing an origin of the
2-dimensional space ---space $(0,0)$. This will be explained shortly.
\begin{definition}[$\scsp$]
An $\scsp$ consists of (1) a finite number of variables ranging over the
universe of points of the 2-dimensional space (henceforth 2D-points); and (2) $\scsp$
constraints on the variables.
\end{definition}
An $\scsp$ constraint is either unary or binary, and either basic or disjunctive.
A basic constraint is
(1) of the form $e(x,y)$, $e$ being equality,
(2) of the qualitative form $\lbound r\rbound (x,y)$ or
$\lbound r\rbound (x)$, depending on whether it is binary or unary, $r$ being a
cone-shaped or projection-based atomic relation of cardinal directions other than
equality, $\imath ,\jmath\in\{0,1\}$,
or (3) of the quantitative form $\lbound\alpha ,\beta\rbound (x,y)$ (binary) or
$\lbound\alpha ,\beta\rbound (x)$ (unary), with $\alpha ,\beta\in [0,2\pi )$,
$(\beta -\alpha )\in [0,\pi ]$, $\imath ,\jmath\in\{0,1\}$.
$\langle ^0$ and $\langle ^1$ stand, respectively, for the left open bracket
( and the left close bracket [. Similarly,
$\rangle ^0$ and $\rangle ^1$ stand, respectively, for the right open bracket )
and the right close bracket ]. A graphical illustration of a quantitative basic
constraint is provided in Figure \ref{basicconstraint}.

\subsection{Translating a qualitative basic constraint into a quantitative basic
constraint}
A qualitative basic relation $\lbound r\rbound$ includes (resp. excludes)
its lower bound, which is a half-line, if $\imath =1$ (resp. $\imath =0$); it
includes (resp. excludes)
its upper bound, which is also a half-line, if $\jmath =1$ (resp. $\jmath =0$).
This means that the version of Frank's relations of cardinal directions we are
using is such that, the region associated with an atom (see Figure
\ref{sectors}) may include both, one or none of its delimiting half-lines.

We remind the reader that we have chosen our Cartesian system of coordinates,
$(O,x'x,y'y)$, in such a way that, on the one hand, the $x$-axis $x'x$ is
parallel to, and has the same orientation as the \mbox{West-East} directed
line of Frank's \mbox{projection-based} model, and, on the other hand, the
\mbox{$y$-axis} $y'y$ is parallel to, and has the same orientation as the
\mbox{South-North} directed line of the same model. The $x$-axis $x'x$ is the
origin of angles, and the anticlockwise orientation is the positive
orientation for angles. Given that we use the set $[0,2\pi )$ as the universe
of angles, if two angles $\alpha$ and $\beta$ are so that $\alpha >\beta$, the
interval $\lbound\alpha ,\beta\rbound$ will represent the union
$\lbound\alpha ,2\pi )\cup [0,\beta\rbound$. Furthermore, given any
$\alpha ,\beta\in [0,2\pi )$, the difference $\beta -\alpha$ will measure the (angular) distance of $\beta$ relative to $\alpha$: the length, in radians, of the anticlockwise ``walk'' from $\alpha$ to $\beta$ (this is, in other words, the size of anticlockwise sector determined by $[\alpha ,\beta ]$).

The atom $\northcs$, for instance, is bounded by the lines whose angular
distances from the $x$-axis are $\frac{3\pi}{8}$, for the lower bound, and $\frac{5\pi}{8}$, for the
upper bound (see Figure \ref{sectors}(left)). The qualitative basic constraint
$\lbound\northcs\rbound (x,y)$ is thus equivalent to the quantitative basic
constraint $\lbound\frac{3\pi}{8},\frac{5\pi}{8}\rbound (x,y)$.
The atom $\northeastpb$ is associated with the region bounded by angles $0$ and
$\frac{\pi}{2}$: the constraint $\lbound\northeastpb\rbound (x,y)$ can thus equivalently be
represented as the quantitative basic constraint $\lbound 0,\frac{\pi}{2}\rbound (x,y)$.
In a similar way, $\lbound\northpb\rbound (x,y)$ is equivalent to
$\lbound\frac{\pi}{2},\frac{\pi}{2}\rbound (x,y)$.\footnote{The reader should be convinced
that the constraint $\lbound\frac{\pi}{2},\frac{\pi}{2}\rbound (x,y)$ is consistent $\iaff$
$\imath =\jmath =1$. A similar remark applies to the other qualitative basic constraints
built from a 1-dimensional projection-based atom ($\eastpb$, $\westpb$ and $\southpb$).}
The other qualitative basic constraints, either cone-shaped or projection-based, are translated in a
similar way. The situation is summarised in the table below.

\begin{footnotesize}
$
\begin{array}{|l|l||l|l|}  \hline
\mbox{$\cdalgcs$ basic constraint}
	&\mbox{Translation}
	&\mbox{$\cdalgpb$ basic constraint}
	&\mbox{Translation}\\  \hline\hline
\lbound\northcs\rbound (x,y)
	&\lbound\frac{3\pi}{8},\frac{5\pi}{8}\rbound (x,y)
	&\lbound\northpb\rbound (x,y)
	&\lbound\frac{\pi}{2},\frac{\pi}{2}\rbound\rbound (x,y)\\  \hline
\lbound\northeastcs\rbound (x,y)
	&\lbound\frac{\pi}{8},\frac{3\pi}{8}\rbound (x,y)
	&\lbound\northeastpb\rbound (x,y)
	&\lbound0,\frac{\pi}{2}\rbound (x,y)\\  \hline
\lbound\eastcs\rbound (x,y)
	&\lbound\frac{15\pi}{8},\frac{\pi}{8}\rbound (x,y)
	&\lbound\eastpb\rbound (x,y)
	&\lbound 0,0\rbound (x,y)\\  \hline
\lbound\southeastcs\rbound (x,y)
	&\lbound\frac{13\pi}{8},\frac{15\pi}{8}\rbound (x,y)
	&\lbound\southeastpb\rbound (x,y)
	&\lbound\frac{3\pi}{2},0\rbound (x,y)\\  \hline
\lbound\southcs\rbound (x,y)
	&\lbound\frac{11\pi}{8},\frac{13\pi}{8}\rbound (x,y)
	&\lbound\southpb\rbound (x,y)
	&\lbound\frac{3\pi}{2},\frac{3\pi}{2}\rbound (x,y)\\  \hline
\lbound\southwestcs\rbound (x,y)
	&\lbound\frac{9\pi}{8},\frac{11\pi}{8}\rbound (x,y)
	&\lbound\southwestpb\rbound (x,y)
	&\lbound\pi ,\frac{3\pi}{2}\rbound (x,y)\\  \hline
\lbound\westcs\rbound (x,y)
	&\lbound\frac{7\pi}{8},\frac{9\pi}{8}\rbound (x,y)
	&\lbound\westpb\rbound (x,y)
	&\lbound\pi ,\pi\rbound (x,y)\\  \hline
\lbound\northwestcs\rbound (x,y)
	&\lbound\frac{5\pi}{8},\frac{7\pi}{8}\rbound (x,y)
	&\lbound\northwestpb\rbound (x,y)
	&\lbound\frac{\pi}{2},\pi\rbound (x,y)\\  \hline
\end{array}
$
\end{footnotesize}

Thus we can, without loss of generality, suppose that a basic constraint is of the form $e(x,y)$, or of the quantitative form $\lbound\alpha ,\beta\rbound (x,y)$.
A disjunctive constraint is of the form
$[S_1
			   \vee\cdots\vee
			       S_n](x,y)$
(binary) or
$[S_1
			   \vee\cdots\vee
			       S_n](x)$
(unary),
with $S_k(x,y)$ and
$S_k(x)$,
$k=1\ldots n$, being basic constraints as described above: in the binary case, the
meaning of such a disjunctive constraint is that, for some $k=1\ldots n$,
$S_k(x,y)$ holds;
similarly, in the unary case, the
meaning is that, for some $k=1\ldots n$,
$S_k(x)$ holds.
A unary constraint $R(x)$ may 
be seen as a special binary constraint if we consider an origin of the 
World (space $(0,0)$), represented, say, by a variable $x_0$: $R(x)$ is
then equivalent to $R(x,x_0)$. Unless explicitly stated otherwise, we
assume, in the rest of the paper, that the constraints of an $\scsp$
are all binary.

An $\scsp$ constraint, $R(x,y)$, is convex if, given an instantiation
$y=a$ of $y$, the set of points $x$ satisfying $R(x,a)$ is a convex
subset of the plane. A universal $\scsp$ constraint is an $\scsp$
constraint of the form $[0,2\pi )(x,y)$: the knowledge consisting
of such a constraint is equivalent to ``no knowledge'', i.e., any
instantiation $(a,b)$ of the pair $(x,y)$ satisfies it. A universal
constraint is also a convex constraint. A convex $\scsp$
is an $\scsp$ of which all the constraints are convex.
Given its similarity to an $\stp$ (Simple Temporal Problem)
\cite{Dechter91a}, we refer to a convex $\scsp$ as an $\ssp$ (Simple
Spatial Problem). An $\scsp$ is basic if, for all pairs $(x,y)$ of
variables, the $\scsp$ includes a basic constraint of the form
$R(x,y)$ or $R(y,x)$. We refer to a basic $\scsp$ as a $\bsp$ (Basic
Spatial Problem).

The standard path consistency procedure for binary CSPs is guided by three algebraic
operations, the converse of a constraint, the composition of two constraints, and
the intersection of two constraints. These are defined below for $\scsps$.
\subsection{The converse of an $\scsp$ constraint}
The converse of an $\scsp$ relation $R$ is the $\scsp$ relation $R^\smile$ such
that, for all $x$, $y$, $R(x,y)$ $\iaff$ $R^\smile (y,x)$. We refer to the
constraint $R^\smile (y,x)$ as the converse of the constraint $R(x,y)$. The
converse of $e(x,y)$ is clearly $e(y,x)$. The converse of an $\scsp$ quantitative
basic relation $\lbound\alpha ,\beta\rbound (x,y)$ is the $\scsp$ quantitative basic
relation $\lbound\alpha +\pi ,\beta +\pi\rbound (y,x)$, which can be explained by
the simple fact that, given any instantiation $(x,y)=(a,b)$ of the pair $(x,y)$
satisfying the constraint $\lbound\alpha ,\beta\rbound (x,y)$, the angle of the
$x$-axis with the directed line $(ba)$ is obtained by adding $\pi$ to the angle of
the $x$-axis with the directed line $(ab)$.
\subsection{The composition of two $\scsp$ constraints}
The composition of two $\scsp$ relations $R$ and $S$, $R\circ S$, is the most
specific relation $T$ such that, for all $x$, $y$, $z$, if $R(x,y)$ and
$S(y,z)$ then $T(x,z)$. We refer to the constraint $T(x,z)$ as the composition
of the constraints $R(x,y)$ and $S(y,z)$.

We describe how to compute the composition of two basic constraints, from which
derives the composition of two general $\scsp$ constraints.\footnote{The reader
should keep in mind that a quantitative basic constraint
$\lbound\alpha ,\beta\rbound (x,y)$ is so that, the difference $\beta -\alpha$
belongs to $[0,\pi ]$.} Clearly, $e\circ R=R\circ e=R$, for all
$\scsp$ relation $R$. For the general case,
let $R=\langle ^{\imath _1}\alpha _1,\beta _1\rangle ^{\jmath _1}$ and
$S=\langle ^{\imath _2}\alpha _2,\beta _2\rangle ^{\jmath _2}$. The result here
is that, if $\beta _1<\alpha _2<\beta _1+\pi$ and $\beta _2>\beta _1+\pi$, then
$R\circ S$ is the universal relation $[0,2\pi )$, which means that, in such a
case, given the knowledge $R(x,y)$ and $S(y,z)$, no knowledge can be inferred
on the extreme variables $x$ and $z$. Otherwise, $R\circ S$ is obtainable in very much the same way as the composition of two (convex) intervals of the real line (cyclicity of the universe $[0,2\pi )$ of angles is a bit tedious but manageable). Basically, the result is $\lbound\alpha ,\beta\rbound$, where $\alpha$ is the minimum, in a certain sense, of $\alpha _1$ and $\alpha _2$, $\beta$ is the maximum of $\beta _1$ and $\beta _2$, $\imath$ is the logical AND of $\imath _1$ and $\imath _2$, and $\jmath$ is the logical AND of $\jmath _1$ and $\jmath _2$, 0 and 1 being interpreted as FALSE and TRUE, respectively ---left to the reader.
\subsection{The intersection of two $\scsp$ constraints}
The intersection of two $\scsp$ relations $R$ and $S$, $R\cap S$, is the $\scsp$
relation $T$ such that, for all $x,y$, the conjunction $R(x,y)\wedge S(x,y)$ is
equivalent to $T(x,y)$. We refer to the constraint $T(x,y)$ as the intersection
of the constraints $R(x,y)$ and $S(x,y)$. Clearly, we have $R\cap S=S\cap R$
(commutativity).

We describe how to compute the intersection of two basic constraints, from which derives the intersection of two general $\scsp$ constraints. $e\cap e=e$; $e\cap\lbound\alpha ,\beta\rbound =e$ if
$\imath =\jmath =1$; and $e\cap\lbound\alpha ,\beta\rbound =\emptyset$ if
$\imath =0$ or $\jmath =0$.
$\langle ^{\imath _1}\alpha _1,\beta _1\rangle ^{\jmath _1}\cap
\langle ^{\imath _2}\alpha _2,\beta _2\rangle ^{\jmath _2}=[\alpha _1,\alpha _1]\cup [\beta _1,\beta _1]$ if $\alpha _1=\beta _2=\beta _1-\pi =\alpha _2-\pi$ and $\imath _1=\jmath _1=\imath _2=\jmath _2=1$; otherwise, $\langle ^{\imath _1}\alpha _1,\beta _1\rangle ^{\jmath _1}\cap
\langle ^{\imath _2}\alpha _2,\beta _2\rangle ^{\jmath _2}$ is obtainable, again, in very much the same way as the intersection of two (convex) intervals of the real line ---left to the reader.
\subsection{Translating an $\scsp$ constraint into a conjunction of linear inequalities}
We now provide a translation of a quantitative basic
constraint into (a conjunction of) linear inequalities. We will then be able to
translate any $\ssp$ (thus, any $\bsp$) into a conjunction of linear inequalities, and
solve it with the well-known Simplex algorithm. Constraint propagation, based on the
algebraic operations we have defined, and the Simplex can be combined in a solution
search algorithm for general $\scsps$: constraint
propagation will be used at the internal nodes of the search space, as a filtering
procedure, and the Simplex at the level of the leaves, as a completeness-guaranteeing
procedure (the $\scsp$ at the level of a leaf is a path-consistent $\ssp$, but since we
know nothing about completeness of path-consistency for $\ssps$, we need to translate
into linear inequalities and solve with the Simplex).

Given a point $X$ of the plane, we denote by $(x_X,y_X)$ its coordinates. The
translation of $e(X,Y)$ is obvious:
$x_X-x_Y\leq 0\wedge x_Y-x_X\leq 0\wedge y_X-y_Y\leq 0\wedge y_Y-y_X\leq 0$. For the
translation of the quantitative basic constraint
$\lbound\alpha ,\beta\rbound (X,Y)$, we consider:
the left half-plane (open if $\imath =0$, close otherwise) delimited by the directed
line through $Y$, whose angular distance from the $x$-axis is $\alpha$;
the right half-plane  (open if $\jmath =0$, close otherwise) delimited by the directed
line through $Y$, whose angular distance from the $x$-axis is $\beta$; and
the close right half-plane delimited by the directed line through $Y$, whose angular
distance from the $x$-axis is $\alpha +\frac{\pi}{2}$ (see Figure
\ref{basicconstraint} for details).
We denote the three half-planes by $lhp(Y,\alpha ,\imath)$, $rhp(Y,\beta ,\jmath)$ and
$\crhp(Y,\alpha +\frac{\pi}{2})$, respectively. It is now easy to see that the
constraint $\lbound\alpha ,\beta\rbound (X,Y)$ is equivalent to $X\in lhp(Y,\alpha ,\imath)\cap rhp(Y,\beta ,\jmath)$, if $\alpha\not =\beta$; and to $X\in lhp(Y,\alpha ,\imath)\cap rhp(Y,\beta ,\jmath)\cap\crhp(Y,\alpha +\frac{\pi}{2})$, if $\alpha =\beta$.

Thus, all we need is to show how to represent with
a linear inequality each of the following assertions on two points $X$ and $Y$ of the
plane:
\begin{enumerate}
  \item[A1] $X$ lies within the open left half-plane delimited by the directed line through
	$Y$, whose angular distance from the $x$-axis is $\alpha$.
  \item[A2] $X$ lies within the close left half-plane delimited by the directed line through
	$Y$, whose angular distance from the $x$-axis is $\alpha$.
  \item[A3] $X$ lies within the open right half-plane delimited by the directed line through
	$Y$, whose angular distance from the $x$-axis is $\alpha$.
  \item[A4] $X$ lies within the close right half-plane delimited by the directed line through
	$Y$, whose angular distance from the $x$-axis is $\alpha$.
\end{enumerate}
We refer to assertions A1, A2, A3 and A4 as $X\in\olhp (Y,\alpha )$, $X\in\clhp (Y,\alpha )$,
$X\in\orhp (Y,\alpha )$ and $X\in\crhp (Y,\alpha )$, respectively. We refer to the line
through $Y$, whose angular distance from the $x$-axis is $\alpha$, as $D$. We consider eight
cases:
$\alpha =0$,
$0<\alpha <\frac{\pi}{2}$,
$\alpha =\frac{\pi}{2}$,
$\frac{\pi}{2}<\alpha <\pi$,
$\alpha =\pi$,
$\pi <\alpha <\frac{3\pi}{2}$,
$\alpha =\frac{3\pi}{2}$,
$\frac{3\pi}{2}<\alpha <2\pi$.
Since, for all $\alpha$ such that $\pi\leq\alpha <2\pi$ (equivalent to $0\leq\alpha -\pi<\pi$), we have
\begin{enumerate}
  \item $X\in\olhp (Y,\alpha )$ $\iaff$ $X\in\orhp (Y,\alpha -\pi )$,
  \item $X\in\clhp (Y,\alpha )$ $\iaff$ $X\in\crhp (Y,\alpha -\pi )$,
  \item $X\in\orhp (Y,\alpha )$ $\iaff$ $X\in\olhp (Y,\alpha -\pi )$, and
  \item $X\in\crhp (Y,\alpha )$ $\iaff$ $X\in\clhp (Y,\alpha -\pi )$.
\end{enumerate}
we can restrict the study to the first four cases,
$\alpha =0$,
$0<\alpha <\frac{\pi}{2}$,
$\alpha =\frac{\pi}{2}$, and
$\frac{\pi}{2}<\alpha <\pi$.
The result is given by the table below, where, given an angle $\alpha$, $tg\alpha$ denotes the tangent of $\alpha$.
\begin{center}
\begin{scriptsize}
$
\begin{array}{|l||l|l|l|l|}  \hline
	&\alpha =0	&0<\alpha <\frac{\pi}{2}	&\alpha =\frac{\pi}{2}	&\frac{\pi}{2}<\alpha <\pi\\  \hline\hline
X\in\olhp (Y,\alpha )&y_X>y_Y&y_X-y_Y>tg\alpha .(x_X-x_Y)&y_X<y_Y&y_X-y_Y>tg(\pi -\alpha ).(x_X-x_Y)\\  \hline
X\in\clhp (Y,\alpha )&y_X\geq y_Y&y_X-y_Y\geq tg\alpha .(x_X-x_Y)&y_X\leq y_Y&y_X-y_Y\geq tg(\pi -\alpha ).(x_X-x_Y)\\  \hline
X\in\orhp (Y,\alpha )&y_X<y_Y&y_X-y_Y<tg\alpha .(x_X-x_Y)&y_X>y_Y&y_X-y_Y<tg(\pi -\alpha ).(x_X-x_Y)\\  \hline
X\in\crhp (Y,\alpha )&y_X\leq y_Y&y_X-y_Y\leq tg\alpha .(x_X-x_Y)&y_X\geq y_Y&y_X-y_Y\leq tg(\pi -\alpha ).(x_X-x_Y)\\  \hline
\end{array}
$
\end{scriptsize}
\end{center}
\section{Summary}\label{summary}
We have provided a qualitative/quantitative constraint-based, $\tcsp$-like language
for reasoning about relative position of points of the 2-dimensional space. The
language, $\scsps$ (Spatial Constraint Satisfaction Problems), subsumes two existing qualitative calculi of relations of
cardinal directions \cite{Frank92b}, and is particularly suited for applications of
large-scale high-level vision, such as, e.g., satellite-like surveillance of a geographic
area. We have provided all the required tools for the implementation of
the presented work; in particular, the algebraic operations of converse,
intersection and composition, which are needed by path consistency. An adaptation
of a solution search algorithm, such as, e.g., the one in \cite{Ladkin92a} (see also
\cite{Dechter91a}), which would use path consistency as the filtering procedure
during the search, can be used to search for a path consistent $\bsp$ refinement of
an input $\scsp$. But, because we know nothing about completeness of path
consistency for $\bsps$, even when a path consistent $\bsp$ refinement exists, this
does not say anything about consistency of the original $\scsp$. To make the search
complete for $\scsps$, we have proposed to augment it with the Simplex algorithm, by
translating, whenever a leaf of the search space is successfully reached, the
corresponding path consistent $\bsp$ into a conjunction of linear inequalities, which
can be solved with the well-known Simplex algorithm.
\newpage\noindent
\bibliographystyle{plain}
\bibliography{/home/AmarUn/Research/BIBLIO/amar}

\begin{thebibliography}{10}

\bibitem{Allen83b}
J~F Allen.
\newblock Maintaining knowledge about temporal intervals.
\newblock {\em Communications of the Association for Computing Machinery},
  26(11):832--843, 1983.

\bibitem{Baader91a}
F~Baader and P~Hanschke.
\newblock A scheme for integrating concrete domains into concept languages.
\newblock In {\em Proceedings of the 12th International Joint Conference on
  Artificial Intelligence (IJCAI)}, pages 452--457, Sydney, 1991. Morgan
  Kaufmann.

\bibitem{Cohn97b}
A~G Cohn.
\newblock Qualitative spatial representation and reasoning techniques.
\newblock In {\em Proceedings KI: German Annual Conference on Artificial
  Intelligence}, volume 1303 of {\em Lecture Notes in Artificial Intelligence},
  pages 1--30, Freiburg, Germany, 1997. Springer-Verlag.

\bibitem{Cooper89a}
M~C Cooper.
\newblock {An Optimal k-Consistency Algorithm}.
\newblock {\em Artificial Intelligence}, 41(1):89--95, 1989.

\bibitem{DeMorgan1864a}
A~De~Morgan.
\newblock On the syllogism, no. iv, and on the logic of relations.
\newblock {\em Trans. Cambridge Philos. Soc. 10}, pages 331--358, 1864.

\bibitem{DeMorgan66a}
A~De~Morgan.
\newblock {\em On the Syllogism and other Logical Writings}.
\newblock Yale University Press, New Haven, 1966.

\bibitem{Dechter92a}
R~Dechter.
\newblock From local to global consistency.
\newblock {\em Artificial Intelligence}, 55:87--107, 1992.

\bibitem{Dechter91a}
R~Dechter, I~Meiri, and J~Pearl.
\newblock Temporal constraint networks.
\newblock {\em Artificial Intelligence}, 49:61--95, 1991.

\bibitem{Forbus91a}
K~D Forbus, P~Nielsen, and B~Faltings.
\newblock Qualitative spatial reasoning: The clock project.
\newblock {\em Artificial Intelligence}, 51:417--471, 1991.

\bibitem{Frank92b}
A~U Frank.
\newblock Qualitative spatial reasoning about distances and directions in
  geographic space.
\newblock {\em Journal of Visual Languages and Computing}, 3:343--371, 1992.

\bibitem{Freuder78a}
E~C Freuder.
\newblock Synthesizing constraint expressions.
\newblock {\em Communications of the Association for Computing Machinery},
  21:958--966, 1978.

\bibitem{Freuder82a}
E~C Freuder.
\newblock A sufficient condition for backtrack-free search.
\newblock {\em Journal of the Association for Computing Machinery}, 29:24--32,
  1982.

\bibitem{Habel95a}
C~Habel.
\newblock Representing space and time: Discrete, dense or continuous? is that
  the question?
\newblock In C~Eschenbach and W~Heydrich, editors, {\em Parts and Wholes ---
  Integrity and Granularity}, pages 97--107. 1995.

\bibitem{Hayes85b}
P~J Hayes.
\newblock The second naive physics manifesto.
\newblock In J~R Hobbs and R~C Moore, editors, {\em Formal Theories of the
  Commonsense World}, pages 1--36. Ablex, 1985.

\bibitem{Isli02d}
A~Isli.
\newblock {Bridging the gap between modal temporal logic and constraint-based
  QSR as a spatio-temporalisation of ALC(D) with weakly cyclic TBoxes}.
\newblock Technical Report FBI-HH-M-311/02, Fachbereich Informatik,
  Universit\"at Hamburg, 2002.
\newblock Downloadable from
  http://kogs-www.informatik.uni-hamburg.de/~isli/home-Publications-TR.html and
  from http://arXiv.org/abs/cs.AI/0307040.

\bibitem{Ladkin92a}
P~Ladkin and A~Reinefeld.
\newblock Effective {S}olution of qualitative {C}onstraint {P}roblems.
\newblock {\em Artificial Intelligence}, 57:105--124, 1992.

\bibitem{Ligozat98a}
G~Ligozat.
\newblock Reasoning about cardinal {D}irections.
\newblock {\em Journal of Visual Languages and Computing}, 9(1):23--44, 1998.

\bibitem{Lutz03a}
C~Lutz.
\newblock {Combining interval-based temporal reasoning with general TBoxes}.
\newblock {\em Artificial Intelligence}, ...(.):...--..., 2003.
\newblock In Press.

\bibitem{Mackworth77a}
A~K Mackworth.
\newblock Consistency in {N}etworks of {R}elations.
\newblock {\em Artificial Intelligence}, 8:99--118, 1977.

\bibitem{Montanari74a}
U~Montanari.
\newblock Networks of {C}onstraints: fundamental {P}roperties and
  {A}pplications to {P}icture {P}rocessing.
\newblock {\em Information Sciences}, 7:95--132, 1974.

\bibitem{Randell92a}
D~Randell, Z~Cui, and A~Cohn.
\newblock A spatial {L}ogic based on {R}egions and {C}onnection.
\newblock In {\em Proceedings KR-92}, pages 165--176, San Mateo, 1992. Morgan
  Kaufmann.

\end{thebibliography}
\end{document}